\pdfoutput=1

\documentclass[11pt]{article}

\usepackage[]{coling}

\usepackage{times}
\usepackage{latexsym}

\usepackage[T1]{fontenc}

\usepackage[utf8]{inputenc}

\usepackage{microtype}

\usepackage{inconsolata}

\usepackage{graphicx}
\usepackage{amsmath}
\usepackage{algorithm}
\usepackage{algpseudocode}
\usepackage{multirow} 
\usepackage{amssymb}
\title{An Active Learning Framework for Inclusive Generation by Large Language Models}

\author{Sabit Hassan\textsuperscript{$\dagger$} Anthony Sicilia\textsuperscript{$\P$} and Malihe Alikhani\textsuperscript{$\P$}\\
  \textsuperscript{$\dagger$}School of Computing and Information, University of Pittsburgh, Pittsburgh, PA, USA  \\
  \textsuperscript{$\P$}Khoury College of Computer Science, Northeastern University, Boston, MA, USA  \\
  \texttt{sabit.hassan@pitt.edu}, \texttt{\{a.sicilia,m.alikhani\}@northeastern.edu} \\
  }

\begin{document}
\maketitle
\begin{abstract}

Ensuring that Large Language Models (LLMs) generate text representative of diverse sub-populations is essential, particularly when key concepts related to under-represented groups are scarce in the training data. We address this challenge with a novel clustering-based active learning framework, enhanced with knowledge distillation. The proposed framework transforms the intermediate outputs of the learner model, enabling effective active learning for generative tasks for the first time. Integration of clustering and knowledge distillation  yields more representative models without prior knowledge of underlying data distribution and overbearing human efforts. We validate our approach in practice through case studies in counter-narration and style transfer. We construct two new datasets in tandem with model training, showing a performance improvement of 2\%–10\% over baseline models. Our results also show more consistent performance across various data subgroups and increased lexical diversity, underscoring our model's resilience to skewness in available data. Further, our results show that the data acquired via our approach improves the performance of secondary models not involved in the learning loop, showcasing practical utility of the framework.

\end{abstract}

\section{Introduction}

Despite advancements, Large Language Models (LLMs) have been under scrutiny for exhibiting bias toward under-represented groups \cite{nozza-etal-2022-pipelines,baffour-etal-2023-analyzing}. Standard fine-tuning may not mitigate this \cite{zhou-etal-2023-causal} as a random sample drawn from skewed data would mirror biases and the fine-tuned model may fail for under-represented groups. To address this, we introduce a novel {active learning framework for \textbf{generative tasks}, combining clustering and knowledge distillation to yield more inclusive generative models. 

\begin{figure}
\centering
\includegraphics[width=0.99\linewidth]{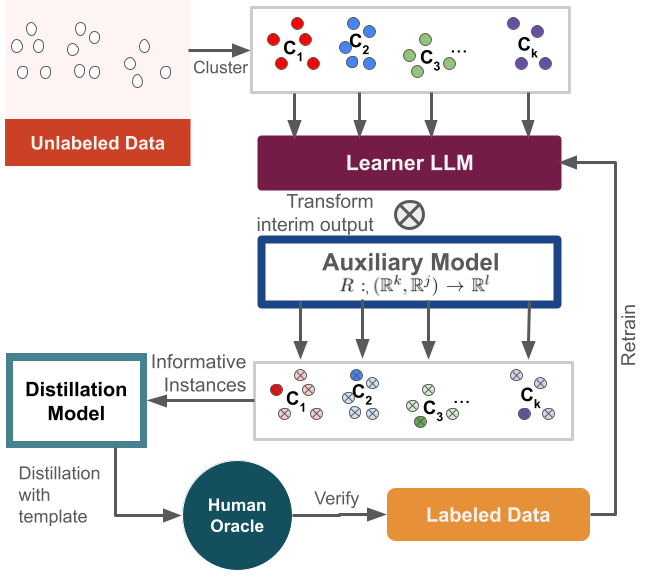}
\caption{The training loop of our framework uses an auxiliary model to transform the interim output of a learner LLM, and selects informative instances from clustered unlabeled data. A distillation model then generates outputs, verified by humans, to iteratively refine the learner LLM.}
\label{fig
}
\end{figure}

Active learning alternates between dataset construction and model training, focusing on the most informative instances \cite{Settles2009ActiveLL}. Unlike classification tasks, active learning for generative tasks faces two key challenges: i) traditional informativeness measures like entropy are ineffective due to vast output spaces, and ii) such measures provide only token-level feedback, complicating full-sequence aggregation. To address these, we propose a new informativeness measure (Section \ref{sec:framework}), computed by an auxiliary model that transforms intermediate output token sequences into a \textbf{1D} latent space, conditioned on a regulated attribute. This measure is not dependent on singular tokens and can be applied to large output space of LLMs.

We advocate for clustering-based active learning in our framework, hypothesizing that informative samples from diverse regions would guard against distributional skewness in real-world data, leading to more inclusive generation. As opposed to traditional bias mitigation approaches that often require extensive post-hoc analysis and rebalancing \cite{han-etal-2022-balancing,sun-etal-2019-mitigating}, 
our method would proactively identify underrepresented groups in data without knowing the underlying distribution beforehand. Additionally, we incorporate knowledge distillation with an external LLM in our framework, leveraging its repository of commonsense and expert knowledge \cite{west-etal-2022-symbolic,hsieh-etal-2023-distilling} to aid the active learner model. The output of the external LLM are verified by human annotators before being passed on to the learner model, reducing necessary human efforts in the active learning paradigm. 



We validate our approach with case studies of counter narration \cite{fanton-etal-2021-human} and style-transfer of offensive text \cite{nogueira-dos-santos-etal-2018-fighting}. We choose these tasks as it is crucial to ensure inclusivity of under-represented groups when addressing harmful text. We construct two novel datasets of \textbf{1K} counter-narrations and \textbf{1K} style transfers, with an emphasis on social acceptability. The datasets contain multiple splits obtained using our framework and baseline methods. This highlights a key difference with prior work as we evaluate the \textbf{practical viability of active learning} beyond mainly simulation-based evaluations \cite{zhang-etal-2022-survey}. Practical evaluation of active learning is challenging as it requires developing multiple splits of data from scratch. While this restricts range of datasets, models and settings compared to simulations, our work answers the critical question of whether active learning can be viable in practice. Our results show that the learner LLMs not only perform well with small datasets but also exhibit greater diversity and lower ratio of errors for under-represented groups (Section \ref{sec:experiments}), suggesting a higher degree of inclusivity. We also validate transferability of acquired data to models outside active learning, showcasing genarliazability. Thus, key contributions in this paper are threefold:
\begin{itemize}
\item Introducing a novel active learning framework for generative tasks, enhanced with clustering and knowledge distillation.
\item Case studies of style-transfer and counter-narration with two publicly available datasets.
\item Demonstrating effectiveness of active learning for generative tasks in practice and its transferability to models outside the learning loop.
\end{itemize}
Our datasets are made publicly available. \footnote{\url{https://github.com/sabithsn/generative-AL}}

\section{Related Work}

\paragraph{Active Learning in NLP} 
Although active learning has been studied for a multitude of NLP tasks \cite{zhang-etal-2022-survey}, almost all existing works target classification scenarios \cite{rotman-reichart-2022-multi,ein-dor-etal-2020-active}. Recent efforts in introducing LLMs with active learning \cite{hassan2024activelearningrobustrepresentative,diao2023active,margatina-etal-2023-active}, assume a fixed set of outputs \textemdash still classification tasks, albeit with LLMs. Active learning for \textit{generative tasks} remains mostly unexplored. \citet{perlitz2023active} show that standard active learning methods for classification do not translate well to generative tasks and call for further investigation. Our work is the first to make significant inroads in active learning for generative tasks.  Different from works that propose LLMs as annotators \cite{xiao-etal-2023-freeal}, our work is the first to utilize knowledge distillation of LLMs, coupled with clustering, for \textit{generative active learning}. Our work is also one of the first to evaluate active learning in practice as opposed to simulations \cite{zhang-etal-2022-survey}.

\paragraph{Countering Distributional Bias} 
NLP models have been under scrutiny for exhibiting bias \cite{Lu2020GenderBI,ahn-oh-2021-mitigating,sap-etal-2019-risk} but reducing such biases has been challenging. First, bias can manifest in a multifaceted way: a model can be biased against gender while also exhibiting linguistic bias \cite{savoldi-etal-2022-morphosyntactic}. Existing approaches often focus on mitigating particular kind of bias such as gender bias \cite{sun-etal-2019-mitigating} by training gender neutral embeddings \cite{zhao-etal-2018-learning} or removing gender direction in embeddding subspace \cite{liang-etal-2020-monolingual}. These methods however, cannot be generalized to address the different types of bias. Another critical challenge is that true distribution of a data is often unknown. As such, existing approaches are often expensive that includes rebalancing \cite{han-etal-2022-balancing} or re-annotating a large amount of data \cite{sun-etal-2019-mitigating}. Our approach offers a generalizable efficent method that does not assume availability of additional data or modification of the model itself.

\paragraph{Counter Narration and Style Transfer} 
Counter narration and style transfer have been suggested as alternative approaches \cite{Mathew2018ThouSN,nogueira-dos-santos-etal-2018-fighting} to filtering-based approaches \cite{ye-etal-2023-multilingual,hassan2020alt} for addressing harmful text.
Prior work in counter-narration primarily address social-media \cite{hassan-alikhani:2023:ijcnlp,Mathew2018ThouSN} or niche-sourced setting \cite{fanton-etal-2021-human,chung-etal-2021-towards}.
Style-transferring offensive text has been limited to social media context \cite{nogueira-dos-santos-etal-2018-fighting,atwell-etal-2022-appdia}.  The output of both of these tasks may not be deemed appropriate when generated by LLMs. Our datasets are the first to consider social acceptability of the generations.



\section{Framework}
\label{sec:framework}

\subsection{Background}
\label{subsec:background}
Most popular active learning paradigms iteratively add informative samples to training data \cite{Settles2009ActiveLL}, offering performance boost over randomly choosing samples. In a randomly chosen sample of skewed real-world data, under-represented groups would remain under-sampled. On the other hand, standard active learning can ignore under-represented groups in data if the learner model is confidently incorrect \cite{hassan2018interactive}. Integrating clustering with active learning can address this challenge and lead to more inclusive models by enforcing informative samples to be chosen from diverse regions in data. In this work, we propose to extend the standard active learning paradigm by applying knowledge distillation to informative samples from each cluster. LLMs like GPT-4 \cite{openai2023gpt4}, which already contain vast amounts of information, can serve to distill knowledge to train other models, reducing exposure to harmful content 
for human annotators. 

This approach, while promising, faces the critical challenge of identifying informative samples. As classification models typically deal with a manageable number of classes, using probabilities in measures such as entropy (Eq. \ref{Entropy}) can gauge informativeness. For generative tasks however, the choices at each timestep match the extensive vocabulary size, typically on the order of $10^4$ \cite{perlitz2023active}, making probability-based measures less reliable. Further, the vocabulary choices are often interchangeable and dependent on previous tokens, which are not taken into account by measures such as entropy. Thus, we propose a novel method to implement active learning for generative tasks next.

\subsection{Active Learning for Generative Tasks}
\paragraph{Preliminaries} We assume there is a large pool of unlabeled dataset \textit{U} but only a small subset of labeled training data \textit{L} can be obtained. \textit{L} is iteratively constructed by querying label for the \textit{most-informative} instance with respect to a learner model $G$. While other active learning scenarios exist \cite{Settles2009ActiveLL}, we focus on pool-based active learning because for many NLP tasks a large amount of unlabeled text data can be obtained from the web. 
The most commonly used measure of informativeness is the uncertainty metric entropy and the instance with highest entropy is defined as: 
\begin{equation}
\label{Entropy}
    x_{E}^* = \underset{x}{argmax} \; - \underset{i}{\sum} P_\theta(y_i|x)logP_\theta(y_i|x)
\end{equation}
In Eq. \ref{Entropy}, i ranges over all possible outputs.  For an instance $x$, the probability of class $y_i$ is denoted as $P_\theta(y_i|x)$. Entropy is higher when probabilities are evenly distributed, indicating greater model uncertainty in class selection. However, as discussed earlier, this approach does not translate well to generative tasks. Thus, we propose a new measure for informativeness next.

\paragraph{Infromativeness Measure for Generative Tasks} To formulate a new measure of informativeness, we introduce the concept of a \textit{regulated attribute $H$}. We define a regulated attribute as a property an LLM is expected to preserve during generation. For instance, when style-transferring offensive text or generating counter-narration, we expect the text to be inoffensive post generation. Thus, inoffensiveness would be the regulated attribute of the task. While we focus on a single regulated attribute in this work due to the nature of the tasks, it can be a weighted combination of multiple attributes. 

We propose training an auxiliary model $R$, to estimate the value of regulated attribute on interim output of the learner model. We can imagine the interim output to be in a \textbf{2-D} space, where each token is a vector. $R$ maps the \textbf{2-D} interim output $G(X_i)$ to a \textbf{1-D} latent space conditioned on regulated attribute $H$. Then, a softmax over this \textbf{1-D} space would inform how well the learner model is adhering to regulated attribute on an instance. While there are no restrictions on $R$'s internal structure (e.g., it can be another pretrained transformer), it is required that the final output layer is a linear layer so that softmax can be applied. Formally, we define the informativeness of a sample as follows:
\begin{equation}
\label{protected-att}
    E_i = Softmax(R(G(x_i), H))
\end{equation}

In the equation, $G(x_i)$ is the LLM generated output on input $x_i$. $R$ is an auxiliary model with a linear output layer that transforms $G(x_i)$ with respect to regulated attribute $H$. Since $R(G(x_i),H)$ is a vector with a single dimension, we can apply softmax over this vector. We replace the entropy portion of Eq. 1 with this measure to obtain most important sample for generative tasks:




\begin{equation}
    x_{E}^* = \underset{x}{argmax} \; - Softmax(R(G(x), H))
\end{equation}

\paragraph{Clustering-based and Knowledge Distillation} 
While the proposed measure is applicable to standard active learning paradigm (Section \ref{sec:experiments}), we advocate for using it in conjunction with clustering as outlined in Section \ref{subsec:background}. The unlabeled data is vectorized and split into m clusters $\{C_1,C_2,...C_m\}$ and most informative samples according to our measure (Eq. 3) are chosen from each cluster. 


Next, we apply knowledge distillation on these samples. We assume we have access to a large language model, $S$, and we want to leverage knowledge distillation of $S$ to assist training of learner model $G$. In our approach, we apply knowledge distillation with the concept of a \textit{template} $T$:
\begin{quote}
    \textit{T(x) : on input x, prompt $S$ to generate $f(x)$ while respecting instruction $I(x)$.} 
\end{quote}
$f(x)$ is the target task, e.g., "style transfer input text from offensive to inoffensive". And $I$ is instruction specific to the task. $I$ can be in different forms depending on the task. For instance, $I(x)$ can be "ensure generated output respects social acceptability". Instead of standard human labeling step, we pass this template to $S$ on instances corresponding to Eq.3 for each cluster $C_i$. The generated content is then verified by a human and added to the training data. The learner model is retrained and this process is repeated until resources runs out. We consolidate our approach in algorithm \ref{alg:nlg-al}. 


\begin{algorithm}
\caption{Knowledge Distilled Clustering-Based Active Learning for Generative Tasks}\label{alg:nlg-al}
\begin{algorithmic}
\State $U,L \gets$ unlabeled data, labeled data
\State $S \gets$ LLM for distillation
\State $G \gets$ bootstrapped model
\State $R \gets$ auxiliary model 
\State $H \gets$ regulated attribute 
\State $B \gets$ labeling budget
\State $N \gets$ annotation batch size
\State $m \gets$ initial number of clusters
\State $V \gets$ vectorize U
\State Cluster $V$ into \{$C_1$, $C_2$, ... $C_m$\}
\While{$B \geq 0$}
    \For{\texttt{i=0,1,...m}}
        \For{\texttt{j=0,1,...$|C_i|$}}
            \State $E_{ij} \gets$ R(G($X_{ij}$), H)
        \EndFor
    
        \State $x_{i}^* \gets \underset{j}{argmax}(E_{ij}$)
        \State $T_{i}^* \gets$ generation template T for $x_{i}^*$
        
        \State $\{(x_{ik}^*, y_{ik}^*)\} \gets$ Distill S with $T_i(x_{i})$
        \State Add $\{(x_{ik}^*, y_{i}^*)\}$ to $L$ 
    \EndFor
    \State $G \gets$ retrain on $L$
    \State $B=B-N$
\EndWhile
\end{algorithmic}
\end{algorithm}
\section{Case Studies}
\label{sec:datasets}

\subsection{Task Definitions}
\label{subsec:task-def}
\begin{table*}[]
    \small
    \centering
    \begin{tabular}{p{4.7cm}|p{4.7cm}|p{4.7cm}}
    \textbf{Sensitive Content} & \textbf{Prior Counter-Narration} & \textbf{Proposed Counter-Narration}\\
    \hline
    \hline
  We should stop immigrants. & If you really wanted to stop the migration, you would destroy our economy, by stopping the flow of people.. \cite{fanton-etal-2021-human}. & immigrants often contribute significantly to our society and economy through their diverse skills and cultures. \\

     \hline
     \hline
        \textbf{Offensive Text} & \textbf{Prior Style-Transfer} & \textbf{Proposed Style-Transfer}\\
        \hline
  you’re just too dumb to see you’re wrong & You just can't see you're wrong. \cite{atwell-etal-2022-appdia} & it seems we have a difference of opinion on this issue \\
    \hline

    \end{tabular}
    \caption{Existing counter-narrative/style-transfer may not be appropriate for LLMs as they can be aggressive. Our proposed approach do not simply counter offensive text or paraphrase them but are also more respectful and polite.}

    \label{tab:prior-work-examples}
\end{table*}

\paragraph{Counter-Narration}
When faced with offensive or sensitive content, an LLM may opt to not respond, potentially allowing harmful ideologies to persist. Counter-narration \cite{fanton-etal-2021-human}, also known as counterspeech \cite{Mathew2018ThouSN}, has been proposed as an alternative to address this issue. \cite{Benesch2014CounteringDS} suggest that counterspeech can be more effective long-term in addressing offensive language. However, these approaches are often discussed within the context of online hate. From an LLM’s safety perspective, overly aggressive responses may alienate users. Therefore, the scope of counter-narration needs to be adjusted to be appropriate for an LLM response in a socially acceptable manner. For social acceptability, an LLM needs to generate counters that are polite and respectful \cite{brown1987politeness}, focusing on non-imposition and the preservation of freedom and desire. We propose utilizing the template to prompt distillation model to adhere to social acceptability and craft more appropriate counter-narration for the learner model. Examples of this approach is illustrated in Table \ref{tab:prior-work-examples} alongside examples from existing works.

\paragraph{Style-Transfer}
Simply censoring model output when it generates offensive language can detract from user experience. Style transfer \textemdash the task of rephrasing text to retain specific stylistic properties without altering the underlying intent \cite{prabhumoye-etal-2018-style}, has been suggested as an alternative method to mitigate offensiveness \cite{nogueira-dos-santos-etal-2018-fighting,atwell-etal-2022-appdia}. However, existing models primarily address social media content and may not ensure a non-aggressive stance essential for LLM safety. We propose adapting style-transfer to not only remove offensiveness but also to ensure it conforms to social expectations that LLMs should not be aggressive toward users. Like counter-narration, we advocate for integrating this social acceptability into the style-transfer process. Table \ref{tab:prior-work-examples} demonstrates how our proposed style-transfer differs from prior work.

\subsection{Datasets}
\label {subsec:datasets}
We contribute two datasets constructed using our framework for counter narration and style-transferring offensive text into inoffensive ones. To construct the datasets, we start with an unlabeled pool of data. From this unlabeled pool of data, instances are chosen either randomly (for baseline) or informative instances added to the train data iteratively following our proposed active learning paradigm. Number of instances in train data is kept small as the goal of active learning is to train models with limited resources. We construct three training splits for each task:

\begin{enumerate}
    \item \textbf{Standard:} Samples are chosen randomly from unlabeled data for fine-tuning.
    \item \textbf{TopN-AL:} N most informative samples are added to training data without clustering.
    \item \textbf{Cluster-AL:} The unlabeled data is clustered into \textit{K} regions.\textit{ N/K} most informative instances are chosen from each region.
\end{enumerate}

Standard fine-tuning with randomly chosen samples serves as baseline. TopN-AL relies solely on informativeness to obtain fine-tuning data but does not utilize clustering. Cluster-AL is the paradigm of our framework, outlined in Algorithm \ref{alg:nlg-al}. All three approaches have access to GPT-4 as distillation model. To evaluate these different paradigms, we choose a randomly chosen subset of the unlabeled data as the test set. The test set is kept fixed with all splits for consistent results.

\paragraph{Counter-Narration Dataset}
We start by creating unlabeled pool of data from the source text of \cite{hassan-alikhani:2023:ijcnlp} and \cite{fanton-etal-2021-human}. \citet{hassan-alikhani:2023:ijcnlp} contains 250 offensive texts and \citet{fanton-etal-2021-human} contains 5K hatespeech targeting women, LGBTQ+, muslims, migrants, disabled and jews. While both of the original datasets contain counter-speech data, we do not use them and create new set of counter-narrations according to Section \ref{subsec:task-def} as counters in \cite{hassan-alikhani:2023:ijcnlp,fanton-etal-2021-human} may not be socially acceptable for use in context of LLM safety (Table \ref{tab:prior-work-examples}).

Utilizing the template with GPT-4 \cite{openai2023gpt4}, we generate counter-narrations of 400 random samples to create the test set. To build the train splits, we use 100 random instances for bootstrapping the learner model. 100 more are added according to definitions of Standard, TopN-AL and Cluster-AL splits. The total count of this dataset is 3X200 + 400 = 1000 pairs.  


\paragraph{Style-Transfer Dataset}
Similar to counter-narration, we start with a set of offensive texts as our unlabeled pool of data. For this, we use offensive-text portion of APPDIA \cite{atwell-etal-2022-appdia} and a subset of the OLID dataset \cite{Zampieri2019PredictingTT}. The APPDIA portion of the data contains 2K offensive comments from Reddit and the OLID portion of the data contains 2K offensive posts from Twitter. Similar to the counter-narration scenario, \citet{atwell-etal-2022-appdia} do contain style-transferred counterparts but we do not use them as they may not be socially acceptable for LLMs.

The construction of the training and test sets follow the same process as counter narration, yielding a total of 3X200 + 400 = 1000 pairs in the dataset. 

\paragraph{Dataset Validation} Two graduate student annotators were hired to verify the counter-narrations and style-transfers distilled from GPT-4. The annotators were provided with definitions of counter-narration and style-transfer from \cite{hassan-alikhani:2023:ijcnlp} and \cite{atwell-etal-2022-appdia} respectively, with the additional constraint that the generations should be socially acceptable (i.e., not aggressive). The distilled generations were found to be reliable in $>98\%$ cases. 
The annotators were hired and paid according to our institution's guidelines. 



\section{Experiments}
\label{sec:experiments}




\subsection{Experiment Setup}
In our experiment setup, we aim to evaluate active learning in practice rather than simulation. To this end, learner models are trained over five iterations  of active learning while constructing the train splits, before evaluation  on the fixed test. The \textit{transfer models} reside outside the active learning setting. The success of our framework would be indicated by not only performance on the two tasks, but also by being inclusive of different groups in data. 
\begin{table*}[]
    \small
    \centering
    \begin{tabular}{p{16cm}}
    \hline
    \textbf{\textit{Offensive Text}} \textit{'All children from the age of 4 will have mandatory lessons on same sex relationships.’ – This is not about Education, this is about willingly confusing and indoctrinating kids... }
    \\
    
    \textbf{Random:} \textcolor{black}{everyone has the right to express their identity.... we should foster understanding and understanding.}\\
    \textbf{TopN-AL:} \textcolor{black}{it's important to remember that every child, regardless of their sexual orientation deserves respect... 
    } \\
    \textbf{Cluster-AL:} \textcolor{black}{everyone has the right to learn about same sex relationships, which also includes children. promoting diversity enhances our society.}\\
    \hline
    \textbf{\textit{Offensive Text}} \textit{Children with Down syndrome are retarded, so more likely to abuse drugs and alcohol.}\\
    \textbf{Random:} \textcolor{black}{actually, down syndrome doesn't identify any specific group, like children, who are more likely to abuse drugs...}\\
    \textbf{TopN-AL:} \textcolor{black}{down syndrome is not a genetic disorder, it's a medical condition that affects individuals with... 
    }\\
    \textbf{Cluster-AL:} \textcolor{black}{people with down syndrome, like anyone, have diverse abilities and temperaments. they deserve respect...}\\
    \hline
    \end{tabular}
    \caption{Examples of counter-narrations generated by different approaches for the learner model Flan-T5. We can observe that randomly choosing samples may result in more failed counter-narrations. Our generations are also more socially acceptable compared to prior work in Table \ref{tab:prior-work-examples}.} 

    \label{tab:cs-examples}
\end{table*}

\setlength{\tabcolsep}{5pt}
\begin{table*}[]
    \centering
    \begin{tabular}{l|c|c||c|c|c|c}
    \hline
     \textbf{Approach} & \multicolumn{2}{c||}{\textbf{FLAN-T5\textsuperscript{\#}}} & \multicolumn{2}{c|}{\textbf{DialoGPT*}} & \multicolumn{2}{c}{
 \textbf{Mistral 7B*}}\\
 \hline
  &   \textbf{CS-Score $\uparrow$}& \textbf{$\downarrow$ Error Var.} & \textbf{CS-Score $\uparrow$}& \textbf{$\downarrow$ Error Var.} &
  \textbf{CS-Score $\uparrow$}& \textbf{$\downarrow$ Error Var.}\\
\hline
  Standard & 65.0 & 0.0101 & 55.5 & 0.0238 & 85.8 & 0.0130\\
  TopN-AL & 70.5 & 0.0061 & 60.3 & 0.0230 & 86.5 & 0.0073\\
  Cluster-AL & \textbf{75.8} & \textbf{0.0049} & \textbf{66.8} & \textbf{0.0155} & \textbf{89.3} & \textbf{0.0018}\\
  \hline
    \end{tabular}
    \caption{CS-Score is the precentange of generated text evaluated to be proper counters. Error Var. denotes variance of error ratios across different targets of hatespeech. A lower value would indicate the errors are less skewed. Our active learning based approach shows efficacy over standard fine-tuning. Clustering-based active learning yields the best results. \textsuperscript{\#} marks the learner model and \textsuperscript{*} shows transferability to other models.
    }
    \label{tab:cs-res}
\end{table*}
\subsection{Models}
\paragraph{Learner Model:} We use FLAN-T5-base \cite{chung2022scaling} as the learner model. FLAN-T5 is an instruction-tuned model with 220 million parameters. We choose FLAN-T5 as learner model as instruction-tuning makes it a capable LLM while its relative smaller size makes it easily deployable. The model is fine-tuned at each iteration of active learning for 10 epochs with learning rate of 3e-5.
\paragraph{Distillation Model:} We choose GPT-4 \cite{openai2023gpt4} for knowledge distillation due its vast knowledge base and advanced generation ability. 

\paragraph{Auxiliary Model:} 
In this work, we use lightweight transformers with attached linear layers as auxiliary models.
For counter-narration, we use a DistillBERT \cite{sanh2020distilbert}, presented in \citet{kim2023robust}, trained to determine if a bot-response is safe in with respect to a human prompt.

For style-transfer, we are only interested in the offensiveness of text post-style transfer, not the input text. We fine-tune a bert-base-cased \cite{Devlin2019BERTPO} on the Jigsaw dataset \footnote{\url{https://www.kaggle.com/c/jigsaw-toxic-comment-classification-\\challenge}}, achieving 93\% macro-averaged F1 score on the Jigsaw test set. 

\paragraph{Clustering:} For vectorizing the data, we use sentence transformer MiniLM-V2 \cite{wang2020minilm}. The vectorized data is clustered using KMeans with default parameters of scikit-learn \footnote{\url{https://scikit-learn.org/stable/modules/generated/sklearn.cluster.KMeans.html}}. Number of clusters is set to 10, similar to prior work in  classification domain \cite{hassan-alikhani-2023-calm}. 

\paragraph{Transfer Models:} Data acquired by the learner model is used to fine-tune a DialoGPT-large \cite{zhang-etal-2020-dialogpt} and a Mistral-7B model \cite{jiang2023mistral}. While Flan-T5 is an encoder-decoder model, DialoGPT is an earlier decoder-only model with 774 million parameters. Mistral-7B on the other hand, is a modern and larger model with 7 billion parameters that utilizes group query attention. These models are chosen to be significantly different from the learner FLAN-T5, and span different generations of LLMs. Transferability across generations would validate ubiquity of our approach.


\subsection{Counter-Narration Results}
\label{sec:counter-narration-results}


     		

\paragraph{Evaluation} Since there is no clear automated way to evaluate counter-narration \cite{hassan-alikhani:2023:ijcnlp}, we conducted human evaluation of 3600 model outputs (400 per model and data split). Table \ref{tab:cs-examples} presents examples of generated counter-narration. We randomized the outputs from the baselines and our proposed model, and tasked two graduate student annotators from earlier (Section \ref{subsec:datasets}) with determining if the generated counter-narration effectively counters the offensive text. A counter-narration is deemed inaccurate if it concurs with the offensive text, strays off-topic, or is incoherent. 
We also calculate the error-ratio-variance with respect to the different target groups present in the original data (e.g., MUSLIM, MIGRANT). This metric is computed by first calculating the error percentage for each group and then the variance across these percentages. A lower variance would suggest that the model is more inclusive by not failing disproportionately for certain groups.

\paragraph{Clustering-based AL yields more effective counters.} From Table \ref{tab:cs-res} we can observe that our proposed approach, Cluster-AL,  outperforms the baseline active learning (without clustering) by \textbf{5.5\%} and random sampling by \textbf{10.8\%} for the learner FLAN-T5. while having access to the same amount of fine-tuning data. 

\paragraph{Clustering-based AL is more inclusive.} We can also observe a substantial reduction in error-ratio-variance; from \textbf{0.01} in standard fine-tuning with random sampling to \textbf{0.0049} for Cluster-AL, which is also lower than baseline active learning \textbf{(0.0061)}. Lower error-ratio-variance coupled with enhanced performance suggests that our method not only produces more accurate counter narrations but also performs more consistently across various groups. This consistency is illustrated in Figure \ref{fig:cs-error}, which displays the error ratio for each class alongside the class's ratio in the original dataset. The initial data pool had few instances targeting Persons of Color (POC) and Disabilities, leading to high error rates with standard fine-tuning with random samples. However, our method reduces error rates for these underrepresented groups, likely due to the clustering step ensuring that active learning selects informative instances from diverse data regions. In skewed data regions (e.g., POC), where random sampling would likely overlook valuable samples, our framework effectively includes them.

\begin{figure}[]
    \centering
    \includegraphics[width=0.99\linewidth]{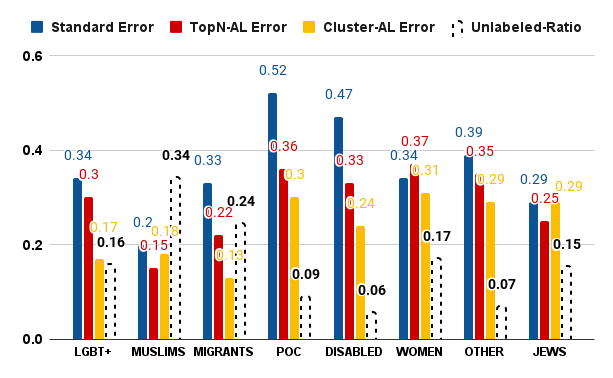}
    \caption{Error ratio of resulting models, along with original data distribution (dashed line). Our clustering-based active learning approach is robust against data distribution skewness.}
    \label{fig:cs-error}
\end{figure}






\subsection{Style-Transfer Results}

\begin{table*}[]
    \small
    \centering
    \begin{tabular}{p{16cm}}
    \hline
    \textbf{\textit{Offensive Text}} \textit{what the f***}\\
    \textbf{Random:} \textcolor{black}{what the f***?}\\
    \textbf{TopN-AL:} \textcolor{black}{i'm sorry, but i can't assist with that} \\
    \textbf{Cluster-AL:} \textcolor{black}{it seems like there might be some misunderstanding here. let's try to clarify this together.}\\
    \hline
    \textbf{\textit{Offensive Text}} \textit{And when they met he looked at her and said I'm in charge in you're my b****}\\
    \textbf{Random:} \textcolor{black}{when they met, he expressed he's in charge of you're my b****.}\\
    \textbf{TopN-AL:} \textcolor{black}{when they met he stated he was in charge in her position and he tried to control the situation}\\
    \textbf{Cluster-AL:} \textcolor{black}{when they came across each other, he looked at her and said he would take the lead}\\
    \hline
    \end{tabular}
    \caption{Examples of style-transfer by the different approaches for the model FLAN-T5. Random sampling may fail to remove offensiveness from complicated samples whereas our proposed approach can do so effectively. Our generations are also more respectful compared to prior work in Table \ref{tab:prior-work-examples}.} 

    \label{tab:st-examples}
\end{table*}

\setlength{\tabcolsep}{5pt}
\begin{table*}[]
    \centering
    \begin{tabular}{l|c|c||c|c|c|c}
    \hline
     \textbf{Approach} & \multicolumn{2}{c||}{\textbf{FLAN-T5\textsuperscript{\#}}} & \multicolumn{2}{c|}{\textbf{DialoGPT*}} & \multicolumn{2}{c}{
 \textbf{Mistral 7B*}}\\
 \hline
  &   \textbf{Safe-Score $\uparrow$}& \textbf{MTLD $\uparrow$} & \textbf{Safe-Score $\uparrow$}& \textbf{MTLD $\uparrow$} &
  \textbf{Safe-Score $\uparrow$}& \textbf{ MTLD $\uparrow$}\\
\hline
Standard & 94.9 & 60.89 & 97.0 & 80.03 & \textbf{99.5} & 124.0\\
TopN-AL & 96.5 & 92.88 & 96.7 & 77.54 & 99.0 & 122.8\\
Cluster-AL & \textbf{97.0} & \textbf{99.81} & \textbf{98.0} & \textbf{82.44} & \textbf{99.5} & \textbf{124.8}\\

  \hline
    \end{tabular}
    \caption{SafeScore is the percentage of style-transferred text predicted to be inoffensive. MTLD measures lexical diversity in generated output. Our proposed approach has more substantial improvement for smaller models FLAN-T5 and DialoGPT. \textsuperscript{\#} marks the learner model and \textsuperscript{*} shows transferability to other models.
    }
    \label{tab:style-res}
\end{table*}

\paragraph{Evaluation} Unlike counter-narration, the offensive counterparts in the original datasets used for style-transfer unlabeled data \cite{atwell-etal-2022-appdia,Zampieri2019PredictingTT} lack tags for offensiveness classes. Therefore, we adopt the Measure of Textual Lexical Diversity (MTLD) \cite{McCarthy2010MTLDVA} to assess diversity. Although MTLD doesn't directly measure bias, a higher MTLD, if other metrics are stable, would suggest that more diverse samples are selected. We also use a BERT-based classifier fine-tuned on the OLID dataset \cite{Zampieri2019PredictingTT} (F1 score 89\%) to compute automated SafeScore \textemdash the percentage of style-transferred text predicted to be inoffensive.

\paragraph{Clustering-based AL yields less offensive style-transfers.} Table \ref{tab:style-res} shows that our approach reduces offensiveness more effectively. For the learner model FLAN-T5, SafeScore increased from \textbf{94.9\%} to \textbf{97\%}. Additionally, the MTLD value significantly increased from \textbf{60.89} to \textbf{99.81}, indicating that our method acquired more diverse samples compared to standard fine-tuning with random sampling. While TopN-AL surpasses the standard approach, Cluster-AL exceeds both.

\paragraph{Clustering-based AL is more robust across data types.} Our analysis of instances deemed offensive post style-transfer in Figure \ref{fig:st-error} reveals that random sampling often fails for under-represented groups like Persons of Color, and overlooks use of uncommon swear words. Both TopN and Cluster-AL approaches reduce offensiveness more effectively, with Cluster-AL showing greater robustness across most data types.
\begin{figure}[]
    \centering
    \includegraphics[width=0.99\linewidth]{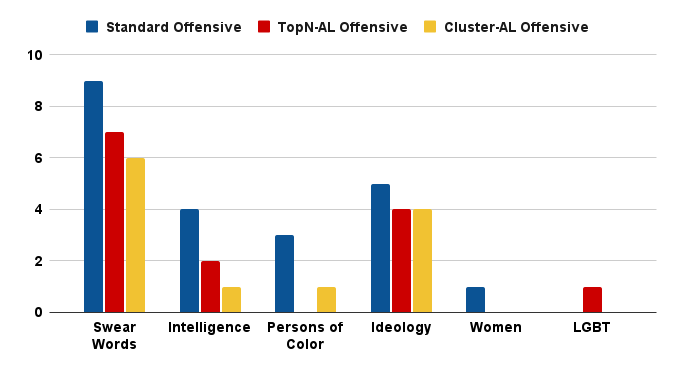}
    \caption{Offensive counts after style transfer. Random sampling leads to higher error rates overall, especially for target groups like persons of color. Cluster-AL achieves the lowest offensiveness overall and for most individual groups.}
    \label{fig:st-error}
\end{figure}
\subsection{Transferability of Active Learning}
 Table \ref{tab:cs-res} and \ref{tab:style-res} demonstrate the transferability of active learning. 
For counter-narration, Cluster-AL shows substantial gains, improving efficacy by \textbf{10\%} over standard fine-tuning for DialoGPT. Although less pronounced, a similar pattern is observed for Mistral 7B, where efficacy increases to \textbf{89.3\%} from \textbf{85.8\%}, and the error ratio variance also decreases significantly for both models (from \textbf{0.0238} to \textbf{0.0155} and from \textbf{0.0130} to \textbf{0.0018}), suggesting that the models are more inclusive of different subgroups in data.

The transferability for style-transfer is less pronounced. While we do see a higher MTLD score for DialoGPT and Mistral 7B, the performance improvement is modest. This can be attributed to the fact that the models were already achieving very high SafeScore. This suggest that the data acquired by active learning is more useful to other models when the task is more challenging.

\section{Conclusion and Future Work}
We presented a novel framework for active learning, making significant inroads toward making active learning viable for generative tasks. By integrating an auxiliary model that transforms interim output, along with clustering and knowledge distillation, our framework effectively generates counter-narrations and style-transfers. The results show that our approach prevents high error rates for under-represented groups and achieves greater lexical diversity. Our approach achieves this without having prior knowledge of the data distribution, highlighting the framework’s potential as a method for more inclusive LLM generation. Creation of two novel datasets and their use across different models validate the practical viability.


Our findings pave the way for further research in expanding active learning to different generative tasks. Future applications could include dialogue systems \cite{sicilia2023isabel}, digital health interventions \cite{wang-etal-2023-medngage} and multimodal generative tasks such as signed language generation \cite{inan-etal-2022-modeling}. Future research could explore different ways of integrating knowledge distillation and clustering and expand on our use of auxiliary models to integrate more complex models. The proposed framework could also be adapted to address specific biases, like behavioral or gender biases.


\section*{Limitations}
We presented a novel framework for active learning for generation and applied the framework in practice by constructing two datasets with multiple splits and training models simultaneously. 
When applying active learning in practice, it is necessary to construct the dataset splits from scratch. Thus, a larger number of simulated experiments and datasets, as seen in prior work, is not feasible in practice. We hope our work can initiate a trend of evaluating active learning in practice so that active learning for generative tasks is adopted broadly in practical applications. 

Additionally, as seen from our results, our proposed approach can draw out under-represented groups from data without knowing underlying distribution. While this improves representation of minority groups, it cannot eliminate the problem of representation completely. Thus, it is still important to monitor behavior of these models before deploying.




\section*{Ethical Considerations}


We presented an annotation efficient approach for drawing out under-represented groups from data. While this comes with better representativeness of minority groups and inclusivity in behavior of generative models, it is important to use the approach responsibly and not change the algorithm to exacerbate biases \textit{against} minority groups.

Since our approach also comes with the benefit of lower involvement of human annotators due to integration of knowledge distillation with clustering-based active learning, we advocate for considering reallocation of saved annotation resources. The saved resources could be used for continued evaluation and training purposes rather than simply reducing human involvement. 

\bibliography{custom}




\end{document}